\documentclass[conference]{IEEEtran}
\IEEEoverridecommandlockouts
\usepackage{cite}
\usepackage{amsmath,amssymb,amsfonts}
\usepackage{graphicx}
\usepackage{textcomp}

\def\BibTeX{{\rm B\kern-.05em{\sc i\kern-.025em b}\kern-.08em
    T\kern-.1667em\lower.7ex\hbox{E}\kern-.125emX}}
    
\graphicspath{{images/}}
\usepackage[table,xcdraw]{xcolor}
\usepackage{algpseudocode,algorithm,algorithmicx}
\usepackage{amsmath}

\usepackage[caption=false,font=footnotesize]{subfig}
\usepackage{tabularx}
\usepackage[hidelinks]{hyperref}
\usepackage{orcidlink}

\newcommand\copyrighttext{%
  \footnotesize \textcopyright 2022 IEEE. Personal use of this material is permitted. Permission from IEEE must be obtained for all other uses, in any current or future media, including reprinting/republishing this material for advertising or promotional purposes, creating new collective works, for resale or redistribution to servers or lists, or reuse of any copyrighted component of this work in other works. IEEE Copyright policy can be found at \textit{\href{https://www.ieee.org/publications/rights/copyright-policy.html}{https://www.ieee.org/publications/rights/copyright-policy.html}}
  }
\newcommand\copyrightnotice{%
\begin{tikzpicture}[remember picture,overlay]
\node[anchor=south,yshift=10pt] at (current page.south) {\fbox{\parbox{\dimexpr\textwidth-\fboxsep-\fboxrule\relax}{\copyrighttext}}};
\end{tikzpicture}%
}

\begin{document}

\title{Agglomerative Hierarchical Clustering with Dynamic Time Warping for Household Load Curve Clustering
\thanks{\hrule\vspace{2pt}\noindent This research has been supported by NSERC under grant RGPIN-2018-06222.}
}

\author{
\IEEEauthorblockN
{
    Fadi AlMahamid , Katarina Grolinger
}
\IEEEauthorblockA
{
    \textit{Department of Electrical and Computer Engineering}\\
    \textit{Western University}\\
    London, Ontario, Canada\\
    Email: \{falmaham, kgroling\}@uwo.ca\\
    ORCID: \orcidlink{0000-0002-6907-7626} 0000-0002-6907-7626, \orcidlink{0000-0003-0062-8212} 0000-0003-0062-8212
}
}

\maketitle
\copyrightnotice

\vspace{-10pt}
\begin{abstract}
Energy companies often implement various demand response (DR) programs to better match electricity demand and supply by offering the consumers incentives to reduce their demand during critical periods. Classifying clients according to their consumption patterns enables targeting specific groups of consumers for DR. Traditional clustering algorithms use standard distance measurement to find the distance between two points. The results produced by clustering algorithms such as K-means, K-medoids, and Gaussian Mixture Models depend on the clustering parameters or initial clusters. In contrast, our methodology uses a shape-based approach that combines Agglomerative Hierarchical Clustering (AHC) with Dynamic Time Warping (DTW) to classify residential households' daily load curves based on their consumption patterns. While DTW seeks the optimal alignment between two load curves, AHC provides a realistic initial clusters center. In this paper, we compare the results with other clustering algorithms such as K-means, K-medoids, and GMM using different distance measures, and we show that AHC using DTW outperformed other clustering algorithms and needed fewer clusters.
\end{abstract}


\section{Introduction} \label{sec:intro}
The estimated number of smart meters in the U.S. and China in 2016 was 166 million \cite{Wang2019}, which is anticipated to increase as International Energy Agency estimates a 90\% rise in power demand by 2040  \cite{Sehovac2020}. The European Union (EU) residential sector represents 31\% of the total energy consumption \cite{Yilmaz2019}. Smart meters enabled energy providers to collect energy usage data more rapidly and accurately, empowering them to analyze and estimate consumer consumption to provide a more cost-effective and reliable Demand Response (DR) \cite{dasgupta2019clustering}. DR balances energy supply and demand by encouraging consumers to use energy when the power supply is ample and affordable. Energy companies apply various marketing strategies to accomplish DR, commonly through incentive programs that target customers based on their consumption patterns to adapt their consumption in response to the change in the prices \cite{Teeraratkul2018}, such as applying different tariffs during the day (peak vs. off-peak hours) \cite{Jin2019}. Moreover, DR actions contribute to reducing CO2 emissions where energy production is considered one of the most significant contributors to global warming producing two-thirds of human-induced greenhouse gas emissions \cite{Sehovac2020}.

Incentive programs aim to segment customers sharing the same consumption, which requires categorizing customers into groups based on their consumption patterns. Customer consumption can be expressed in a load curve representing the consumption pattern in a given period. Office Buildings might share similar load curves because they follow the same daily work routine. In contrast, residential households' load curves vary from one customer to another, and patterns might even differ within the same customer due to changes in device usage \cite{Teeraratkul2018}, which makes it more difficult to segment customers into groups based on their consumption.

Clustering is an unsupervised learning technique that groups similar data instances together \cite{Camastra2015} that can be used to segment customers based on their energy consumption. Different clustering approaches are used to cluster customers based on their consumption \cite{Jin2019,dasgupta2019clustering,kwac2014household}. Some techniques, in addition to the energy consumption data, include customer data such as customer demography and the number of occupants, then apply significance analysis to reduce the number of features (e.g., Principal Component Analysis (PCA)) and then categorize customers into groups \cite{wang2015load, wang2018review}. Other approaches focused on clustering the energy consumption using only daily load curves.

K-means and K-medoids are clustering algorithms based on partitioning: they are used to cluster energy data because of their simplicity \cite{hamerly2003learning}. However, the final results depend on the quality of the selected initial cluster centers, which means they might be drawn to a local optimum \cite{Xu2015, celebi2013comparative}. Other clustering algorithms based on density, such as DBSCAN and OPTICS clustering results, are sensitive to initial parameters and data density distribution \cite{Xu2015}.

Recent literature suggests using a shape-based approach to cluster energy load curves \cite{Teeraratkul2018, dasgupta2019clustering, Jin2019}. Agglomerative Hierarchical Clustering \cite{murtagh1983survey} is a hierarchical-based algorithm suitable for arbitrary shapes data. All clustering algorithms, including shape-based ones, use a similarity-distance to construct clusters, e.g., Euclidean, Manhattan, Cosine, and Dynamic Time Warping. While metrics such as Euclidean and Manhattan calculate distances between pairs of points (or features), Dynamic Time Warping (DTW) helps compare load curves based on shape and detect shifts in consumer consumption patterns.

This paper facilitates shape-based clustering while remedying the challenge of poor initial clusters using Agglomerative Hierarchical Clustering with Dynamic Time Warping (AHC-DTW) for residential load clustering. Agglomerative Hierarchical Clustering is used as a bottom-up approach for cluster formation, whereas the DTW similarity metrics allow AHC to consider shapes as opposed to individual points. The results show that using the AHC-DTW algorithm outperformed different clustering algorithms such as K-means, K-medoids, and Gaussian Mixture Models (GMM), resulting in fewer clusters. Furthermore, this paper shows that using DTW as a similarity distance with AHC improves clustering results compared to other distance measures such as Euclidean, Manhattan, and Cosine.

This paper is organized as the following: Section \ref{sec:background} provides an introduction. Section \ref{sec:related-work} discuss the related work. Where Section \ref{sec:methodology} explains how AHC-DTW used to cluster load profiles. Section \ref{sec:results} discuss the results and Section \ref{sec:conclusion} concludes the paper and discuss future work.

\section{Background} \label{sec:background}
\subsection{Dynamic Time Warping}
\begin{figure}[t]
    \centering
    \includegraphics[width=1\linewidth]{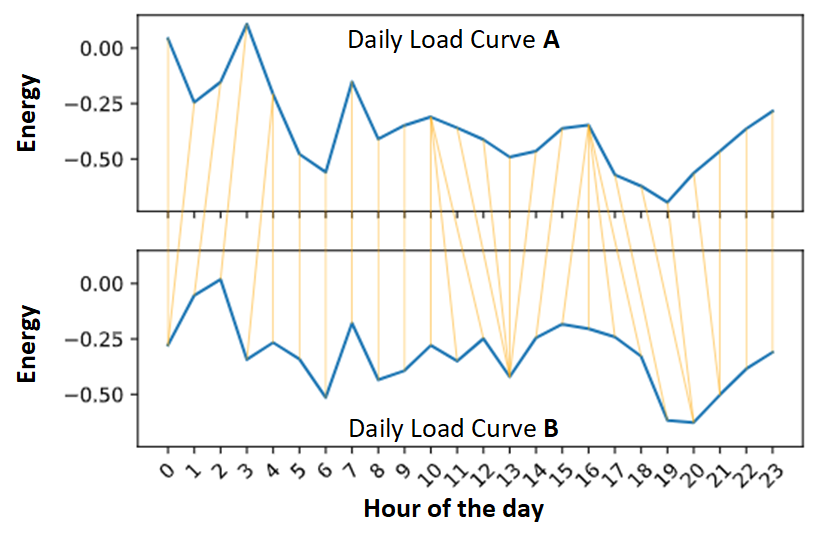}
    \caption{DTW optimal alignment between two daily load curves showing how DTW aligns the points based on similarity}
    \label{fig:dtw-alignment}
\end{figure}

Dynamic time warping (DTW) was used initially to compare speech patterns in speech recognition \cite{Senin2008} and was then extended to identify the optimal alignment between two time-series sequences subject to specific constraints\cite{Muller2007}, as illustrated in Figure \ref{fig:dtw-alignment}. The DTW cost matrix (distance matrix) must be computed to calculate the DTW value between two time-series, quantifying the alignment of the two sequences under specific constraints. The closer the DTW value is to zero, the more similar the shapes of the curves are.

\noindent
Consider two time-series $X$ and $Y$, where $X = \{ x_1,x_2,x_3,...,x_n \}$ and $Y = \{ y_1,y_2,y_3,...,y_m \}$, and $C_\ell$ cost value between two points $x_i$ and $y_j$, where $x_i \in X$, and  $y_j \in Y$. Then, $W_i$ defines a warping path between $X$ and $Y$ as follows: 

\begin{equation}
    \label{eq:w}
    W_i = \sum_{\ell=1}^{L} C_\ell \quad \forall \quad C_{\ell} \in W_i
\end{equation} 

\noindent 
Therefore, for all warping paths $W_\mathbb{P} = \{W_1, W_2,...,W_p\}$, DTW is defined as the follows:

\begin{equation}
    \label{eq:dtw}
    DTW(X,Y) = \min\{W_\mathbb{P}\}
\end{equation}
    
\begin{figure}[t]
    \centering
    \includegraphics[width=1\linewidth]{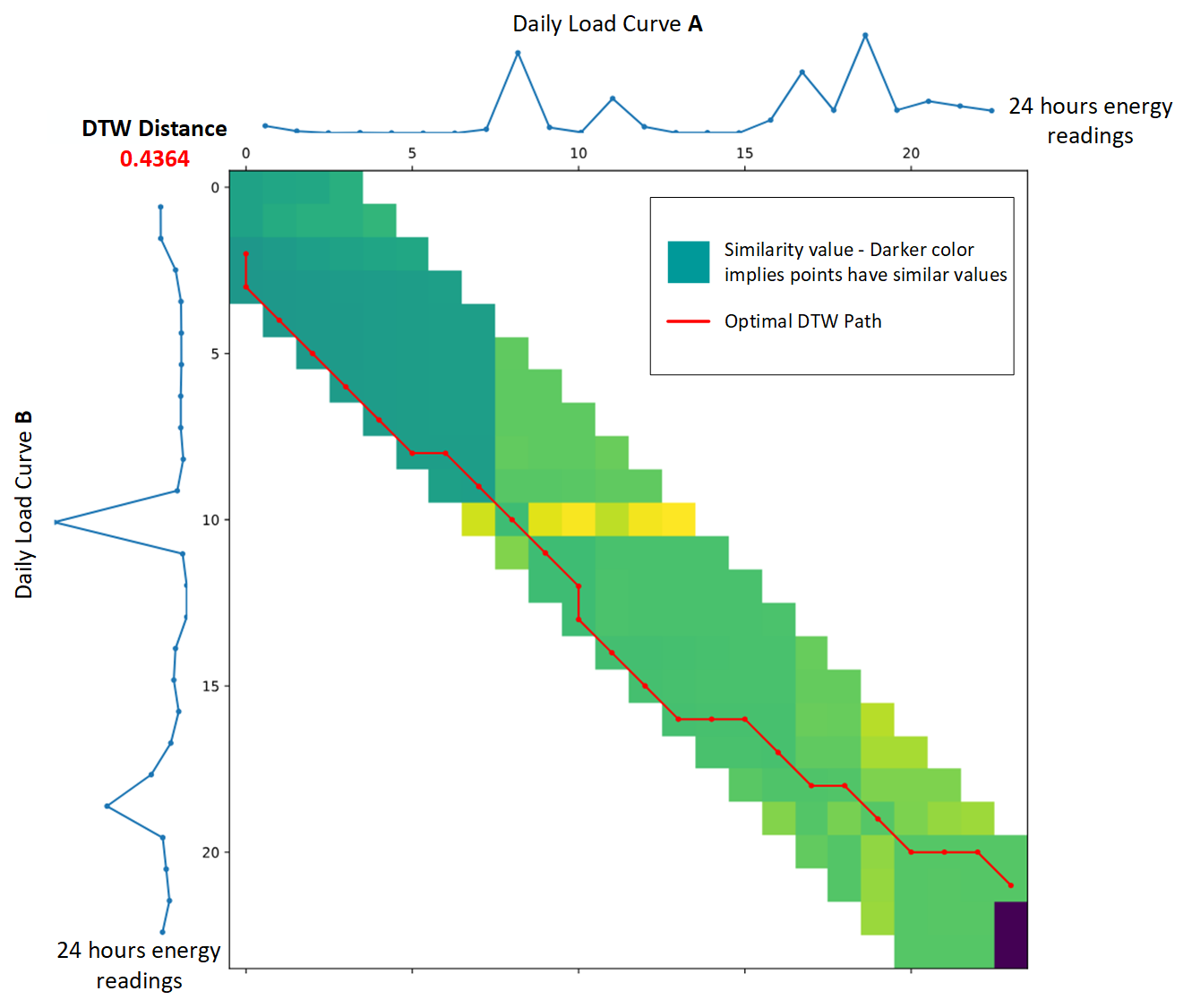}
    \caption{DTW between two load curves showing the optimal warping path in red line using a window size of four}
    \label{fig:dtw-two-load-curves}
\end{figure}

Equation \ref{eq:dtw} defines DTW as the minimum of all warping paths between $X$ and $Y$, where a warping path represents the sum of all distances in a distance-matrix path. The solid red line in Figure \ref{fig:dtw-two-load-curves} shows the minimum warping path between the two load curves, and the sum of all points in that path represents the DTW value.

\begin{figure}[t]
    \centering
    \includegraphics[width=1\linewidth]{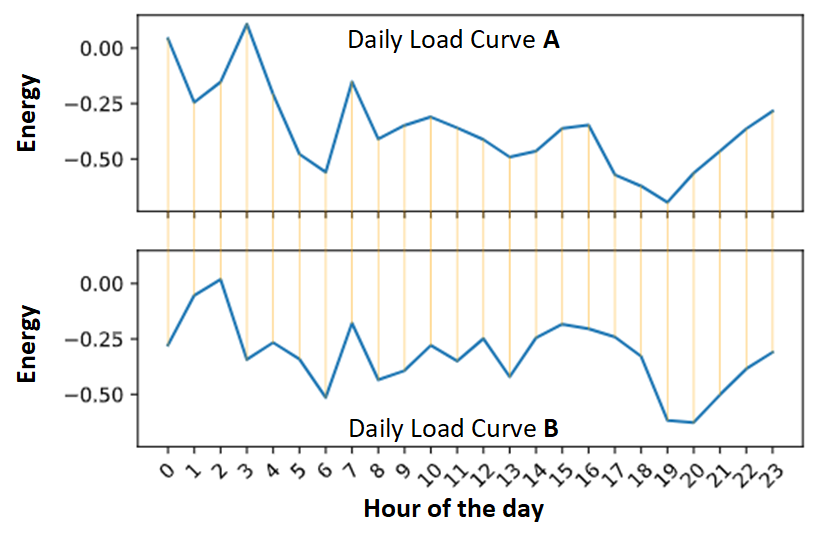}
     \caption{DTW between two daily load curves using the window size equal to one. This alignment is equivalent to using Euclidean distance}
    
    \label{fig:dtw-alignment-w1}
\end{figure}

DTW is capable of capturing the shift in patterns between two time-series using a defined window size that determines the number of proceeding points that need to be considered when aligning a selected point. For example, a DTW with window size of four considers four consecutive points $y_i,y_{i+1},y_{i+2}, \textrm{ and } y_{i+3}$ in the curve $Y$ while aligning a selected point $x_i$ in the curve $X$. On the other hand, a window size of one considers only the corresponding point $(x_i, y_i)$, which is equal to computing the Euclidean distance between two time-series as shown in Figure \ref{fig:dtw-alignment-w1}. 

\subsection{Hierarchical Clustering}
Hierarchical Clustering is a clustering algorithm that iteratively partitions the instances into clusters either using a bottom-up (Agglomerative) or top-down (Divisive) approach \cite{Camastra2015}. Agglomerative Hierarchical Clustering (AHC) initially considers each instance as a cluster, and then the clusters are recursively merged until the desired number of clusters is achieved, or all clusters are merged into one cluster. In contrast, Divisive Hierarchical Clustering (DHC) initially merges all instances into a single cluster, and then the clusters are iteratively divided into sub-clusters until the desired number of clusters is achieved.

AHC first computes the distance matrix, which defines the distances between all the instances to be clustered. Then it merges the two instances with the shortest distance, which requires updating the distance between the newly merged instances and all other remaining instances. There are different methods to update the distance matrix, commonly referred to as the linkage criteria, such as:

\noindent
\textbf{Single-linkage clustering} computes a distance matrix between all existing clusters, then merges the two clusters separated by the shortest distance. The newly merged cluster retains the minimum distance between the merged cluster and the remaining clusters.\\
Let clusters $A$ and $B$ have the shortest distance in the distance matrix, and $K$ represents the remaining clusters. The function $D(A, B)$ defines the distance between two clusters, and $C$ represents the new merged cluster. Then distance values are defined as:

\begin{multline}
    \label{eq:Single-linkage}
    D(C,K_i) = Min \Big\{ D(A,K_i) , D(B,K_i)\Big\} \; \forall \;  K_i \in K         
\end{multline}

\noindent
\textbf{Complete-linkage clustering}, like single-linkage clustering, computes a distance matrix between all existing clusters, then merges the two clusters separated by the shortest distance. However, in complete-linkage clustering, the new merged cluster retains the maximum distance between the merged clusters and the remaining clusters, as explained in Equation \ref{eq:complete-linkage}.
\begin{multline}
    \label{eq:complete-linkage}
    D(C,K_i) = Max \Big\{ D(A,K_i) , D(B,K_i)\Big\} \; \forall \;  K_i \in K        
\end{multline}

\noindent
\textbf{Unweighted average linkage clustering} is also known as Unweighted Pair Group Method with Arithmetic mean (UPGMA). It is similar to other linkage criteria, except when merging the two clusters separated by the shortest distance, it computes the average distance between the merged clusters and remaining clusters as shown in Equation \ref{eq:UPGMA}.

\begin{multline}
    \label{eq:UPGMA}
    D(C,K_i) = AVG \Big\{ D(A,K_i) , D(B,K_i)\Big\} \; \forall \;  K_i \in K       
\end{multline}

\section{Related Work} \label{sec:related-work}
This section focuses on clustering load curves for residential households using a shape-based approach that has been dominant in recent years. Various clustering algorithms have been used with different similarity distances for comparing the similarity between two time series such as Euclidean, Manhattan, Squared Euclidean, Mahalanobis distance, and Dynamic Time Warping. Zhang \textit{et al.} \cite{Zhang2016} used the K-means clustering with Euclidean distance to cluster 24-hour load curves. Jin and Bi \cite{Jin2019} used Affinity Propagation (AP) clustering using DTW as a distance measure, where data points exchange messages between each other until a high-quality set of exemplars and corresponding clusters emanate \cite{Frey2007}. Teeraratkul \textit{et al.} \cite{Teeraratkul2018} used K-medoids replacing Euclidean distance with DTW and produced better quality clusters compared to K-means and Gaussian-based Expectation-Maximization (EM) algorithms. Yilmaz \textit{et al.} \cite{Yilmaz2019} considered two approaches and compared them. The first approach considered the shape of the daily profiles using the K-means with standard Euclidean distance to cluster daily load profiles and tried to optimize the $K$ value using silhouette score \cite{rousseeuw1987silhouettes}, while the second approach used specific features to cluster daily profiles. Zhang \textit{et al.} \cite{zhang2022clustering} applied hierarchical clustering using DTW to cluster load profiles of a \textit{ground source heat pump system} and determined the best cluster number using the sum of squares of errors, then analyzed the pattern of each cluster class.

Dynamic Time Warping (DTW) distance is used in some literature as a replacement for the Euclidean distance \cite{dasgupta2019clustering, Teeraratkul2018, Jin2019} in clustering load profiles. DTW finds the optimal alignment between two time-series compared to Euclidean distance, which measures the distance between two straight points. The optimal alignment is obtained by stretching or compressing the series' segments \cite{Teeraratkul2018}.

Using a shape-based approach to cluster load curves was not only limited to DTW; for example, Dasgupta \textit{et al.} \cite{dasgupta2019clustering} uses elastic shape analysis to cluster and analyze load curves according to their shapes. On the other hand, Eskandarnia \textit{et al.} \cite{eskandarnia2022embedded} developed a framework that uses auto-encoders to perform dimensionality reduction and provide clustering-friendly representations that maintain the original data characteristics before performing clustering using KL divergence.

Our approach is different since it combines Agglomerative Hierarchical Clustering (AHC) with Dynamic Time Warping. While AHC provides realistic cluster centers since it uses actual load curves as initial class centers (prototypes), DTW assists by comparing the curve shapes and finding the optimal alignment between two curves rather than the straight distance and, therefore, detecting shifts in consumption patterns within predefined  sliding-window.

\section{Methodology} \label{sec:methodology}

\begin{figure}[!ht]
    \centerline{
    \includegraphics[width=1\linewidth]{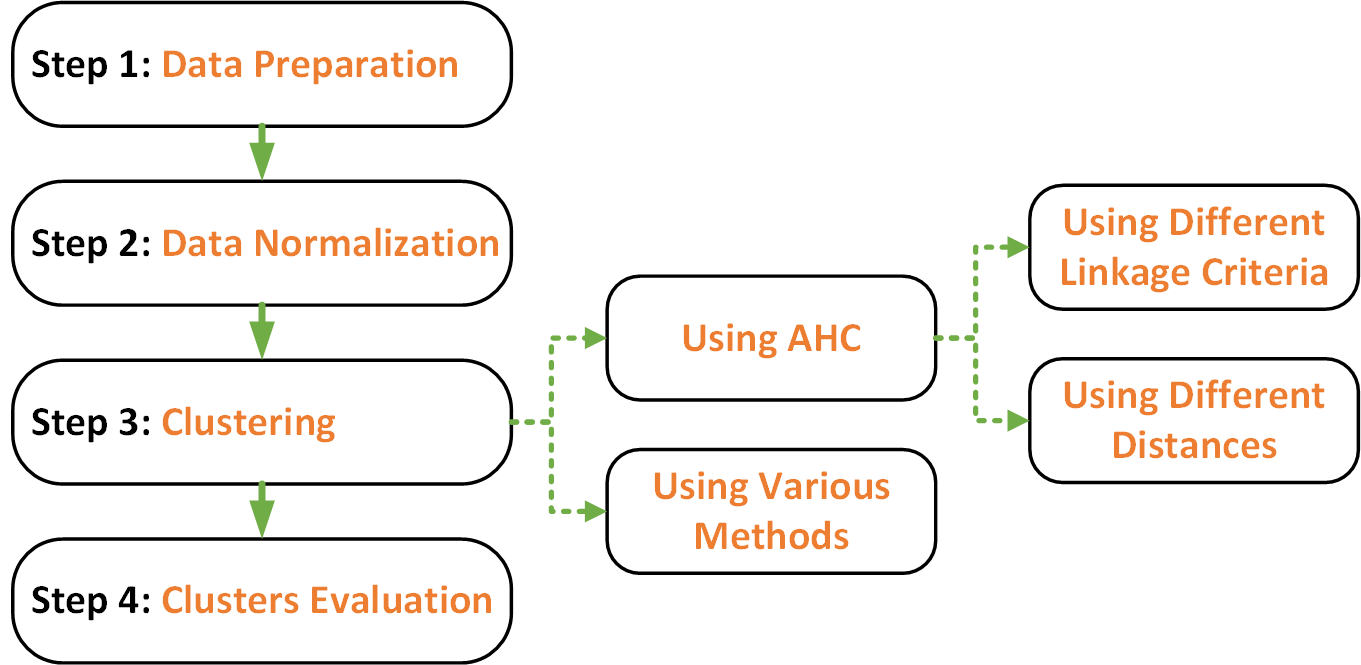}
    }
    \caption{Clustering Methodology performed with various clustering methods, distances, and linkage criteria}
    \label{fig:methodology}
\end{figure}

As seen in Figure \ref{fig:methodology}, the proposed clustering process includes several steps to achieve high-quality clusters and evaluate the findings. The following subsections discuss each step in more detail.

\subsection{Dataset Preparation}
The dataset contains hourly energy consumption for 19 households captured between 2014 and 2016. The households' data provided by London Hydro (London, Ontario, Canada) has no personal information identifying the customers, only generic values to identify the hourly energy consumption reading per household, e.g., \textit{House\#1}.

\begin{figure}[t]
    \centerline{
    \includegraphics[width=.8\linewidth]{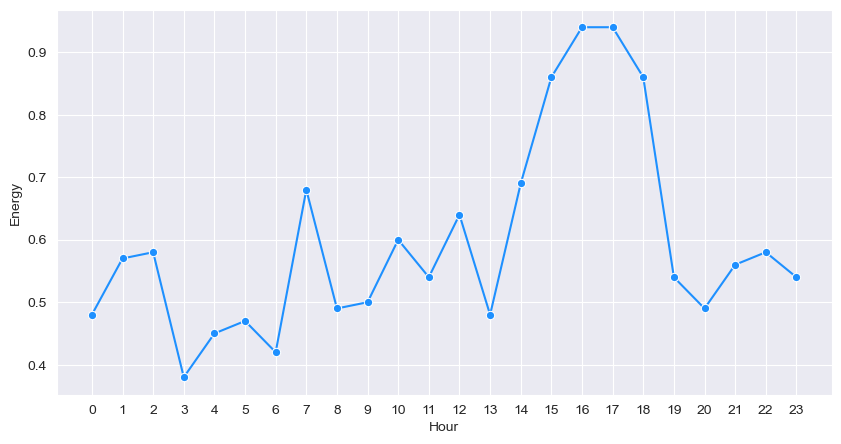}
    }
    \caption{Daily Load Curve (DLC) using 24 hourly energy readings}
    \label{fig:dlc}
\end{figure}

The data were reformatted so each sample (clustering instance) would represent a daily load curve containing 24 hourly energy consumption readings for a particular house during a specific day of the year. Therefore, the household energy consumption during the day is represented using a \textit{Daily Load Curve} (DLC) as shown in Figure \ref{fig:dlc}.

\subsection{Data Normalization}
Data normalization is essential for comparing DLCs shapes using DTW for two reasons: 1) DTW is used to compare the time-series shapes, not the actual values, and 2) the differences in the mean or standard deviation (STD) impact any shape similarity. Therefore, Z-normalization in Equation \ref{eq:z-normalization} is applied to normalize DLC features.

\begin{equation}
    \label{eq:z-normalization}
    Z_i = \frac{X_i - \overline{X}}{\sigma}  
\end{equation}

\noindent

\begin{figure*}[!ht]
	\subfloat[Data Before Normalization]{
		\includegraphics[width=0.5\linewidth]{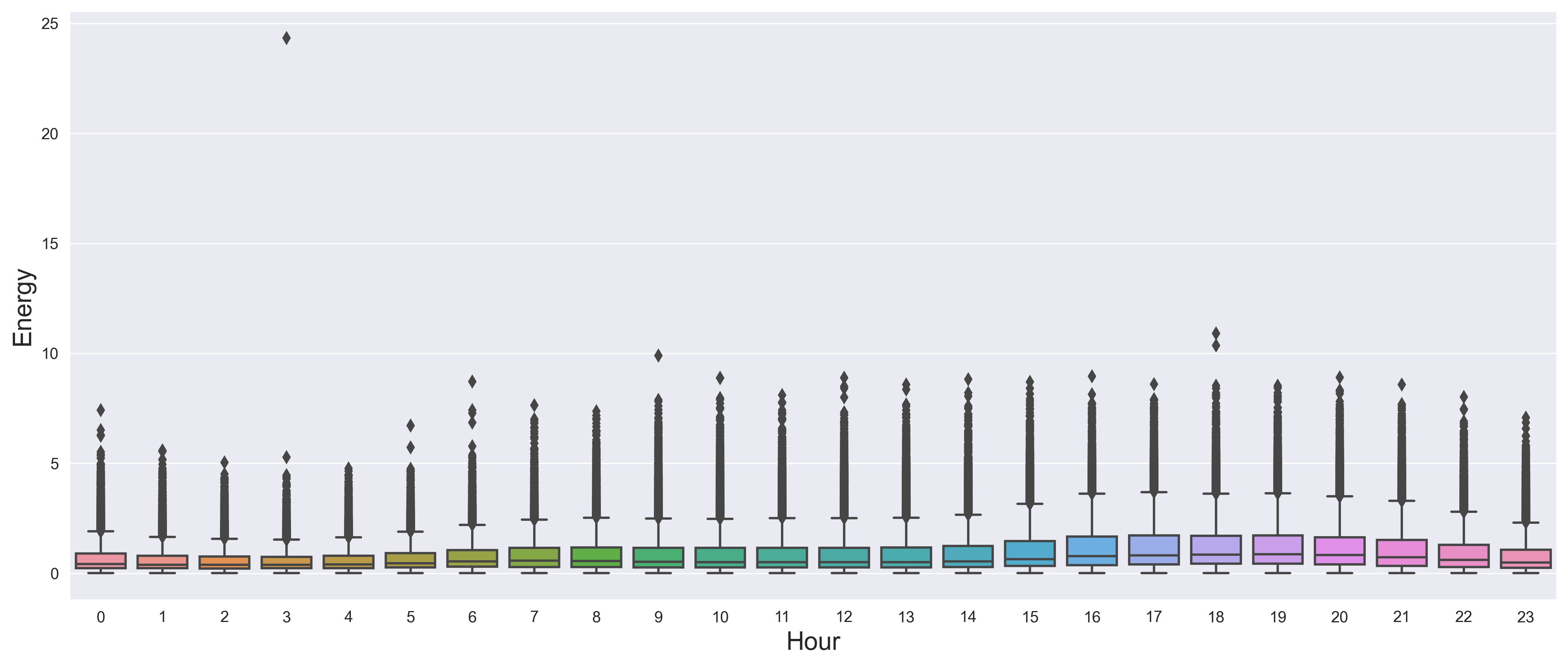}
		\label{fig:data-visual-normal}
	}
	\hfill
	\subfloat[Z-Normalized Data]{
		\includegraphics[width=0.5\linewidth]{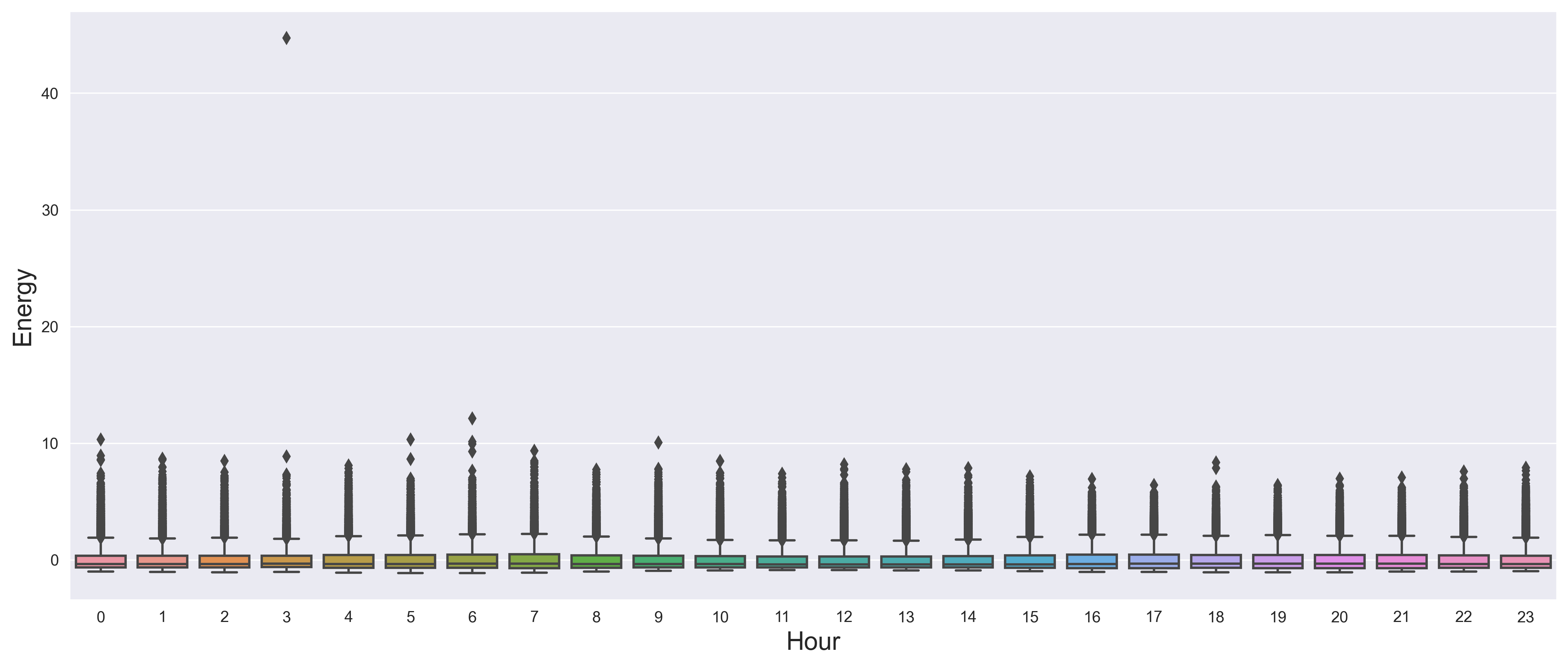}
		\label{fig:data-visual-normalized}
	}
	\caption{Data visualization before and after applying normalization shows that applying normalization reduced the STD and the outliers}
	\label{fig:data-visual}
\end{figure*}

Here $Z_i$ represents the normalized value which is the difference between the actual value $X_i$ and the mean, divided by the STD. Figure \ref{fig:data-visual} shows the data distribution before and after normalization. Figure \ref{fig:data-visual-normal} shows some outliers and a high STD during most of the day before applying normalization. In contrast, Figure \ref{fig:data-visual-normalized} shows that applying normalization reduced the STD and the outliers.

\subsection{Clustering}
DLCs are clustered using different clustering methods AHC-DTW, K-means, K-means++, K-medoids with DTW, and Gaussian Mixture Model with Expectation-Maximization (EM GMM).

AHC was performed utilizing various linking criteria, including complete, average, and single linkage. Furthermore, to determine the influence of DTW on AHC and how it enhances the clustering quality, AHC was performed using several similarity-distances such as DTW, Euclidean, Cosine, and Manhattan.

Despite occasional claims that DTW could add a shape distortion (pinching effect \cite{srivastava2016functional}), \cite{dasgupta2019clustering}; DTW is used here to detect similar energy consumption within four hours period, which would not cause DLC shape distortion while finding the optimal alignment as shown in the experiments. Therefore, DTW is strictly applied with $Window=4$ and was not optimized for better clustering results. Note that the approach still clusters the daily load curves and this $Window=4$ only characterizes the alignment in DTW. The window size of four is an appropriate setting to spot shifts in daily consumption patterns while still not allowing overly high distortions.

\subsection{Clusters Evaluation}
Several clustering evaluation metrics may be used to assess the quality of clustering, including the Silhouette Coefficient \cite{rousseeuw1987silhouettes} and Dunn's Index \cite{dunn1973fuzzy}. The quality of clusters is determined in this study using the ratio of within-cluster distances to between-cluster distances (WCBCR) \cite{tsekouras2008new, Teeraratkul2018}, as denoted in Equation  \ref{eq:evluation-metric} because it takes into account both the similarity of the instances within each cluster (individual cluster quality) and the dissimilarity between clusters (all clusters quality). WCBCR is calculated as follows:

\begin{equation}
    \label{eq:evluation-metric}
    WCBCR = \frac{\sum\limits_{k=1}^{K} \sum\limits_{X \in C_k} d(X,\mu_k)}{\sum\limits_{i\neq j}^{K} d(\mu_i,\mu_j)}
\end{equation}

WCBCR represents the ratio of the sum of the distances between all DLCs ($X \in C_k$) in a cluster ($C_k$) and their corresponding cluster's prototype ($\mu_k$) to the sum between all clusters' prototypes $d(\mu_i,\mu_j) \; \forall \; i\neq j \; \in \; K$. In other words, it measures the ratio between the similarity within clusters to the diversity between clusters, where a smaller value of WCBCR indicates a better clustering quality.

Clustering evaluation is a challenging task since the evaluation metrics depend on the similarity distance used for evaluation, and using different distance similarity measures makes it even harder to compare. Therefore, we decided to use the same similarity distance and metric for clustering evaluation to overcome this challenge, i.e., Euclidean distance.

\section{Results and Discussion}\label{sec:results}
\begin{figure}[t]
    \centerline{
    \includegraphics[width=\linewidth]{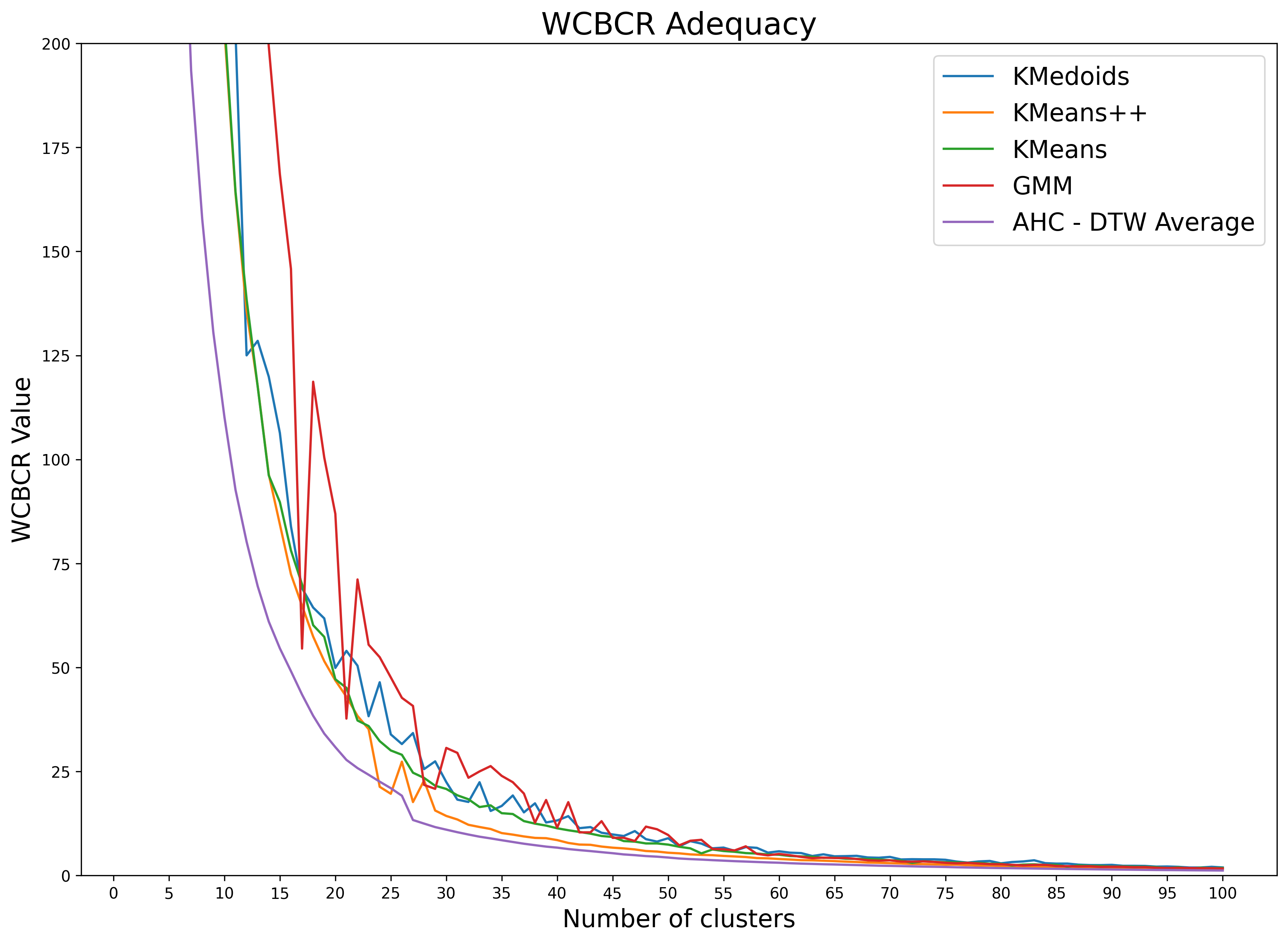}
    }
    \caption{WCBCR results using different clustering methods. The figure shows that using AHC with DTW and average linkage criteria outperformed other clustering methods}
    \label{fig:all-clustering-results}
\end{figure}

WCBCR is computed using various clusters' numbers for AHC-DTW with average linkage criteria and compared to other clustering methods as shown in Figure \ref{fig:all-clustering-results}. The WCBCR value drops as the number of clusters increases for all clustering methods, with AHC-DTW outperforming all other clustering methods by achieving a smaller WCBCR value for the same number of clusters.     

Using the Elbow method that utilizes the curve's elbow to decide on the best number of clusters, we can infer that ACH-DTW outperformed other clustering methods, including K-medoids using DTW. ACH-DTW requires fewer clusters (approximately 27 clusters) while providing a smaller WCBCR value as indicated by the highlighted green cell in Table \ref{tbl:wcbcr-clustering}.

\begin{table}[!ht]
    \caption{WCBCR value using various clustering methods. The green highlighted cell pertains the best WCBCR $13.33$ from AHC-DWT using average linkage criteria with the optimal number of clusters $27$ determined using the elbow method.}
    \centering
    \begin{tabular}{>{\centering}m{0.1\linewidth}
                    >{\centering}m{0.12\linewidth}
                    >{\centering}m{0.1\linewidth}
                    >{\centering}m{0.13\linewidth}
                    >{\centering}m{0.1\linewidth}
                    >{\centering}m{0.15\linewidth}}
        \hline\hline
        \rowcolor{lightgray} Clusters \# & K-medoids & K-means & K-means++ & GMM & AHC-DTW Avg \tabularnewline
        \hline\hline
        20 & 49.89 & 47.16 & 46.89 & 86.93 & 30.85\tabularnewline
        21 & 53.99 & 45.09 & 42.96 & 37.71 & 27.78\tabularnewline
        22 & 50.42 & 37.25 & 38.35 & 71.15 & 25.81\tabularnewline
        23 & 38.28 & 35.92 & 35.18 & 55.47 & 24.21\tabularnewline
        24 & 46.44 & 32.29 & 21.31 & 52.47 & 22.55\tabularnewline
        25 & 33.91 & 30.05 & 19.64 & 47.61 & 20.98\tabularnewline
        26 & 31.59 & 29.01 & 27.35 & 42.70 & 19.17\tabularnewline
        \textbf{27} & 34.23 & 24.70 & 17.65 & 40.75 & \cellcolor[HTML]{03C03C}\textbf{13.33}\tabularnewline
        28 & 25.57 & 23.45 & 22.82 & 21.77 & 12.47\tabularnewline
        29 & 27.43 & 21.57 & 15.61 & 20.82 & 11.63\tabularnewline
        30 & 22.46 & 20.80 & 14.31 & 30.66 & 11.01\tabularnewline
        \hline
    \end{tabular}
    \label{tbl:wcbcr-clustering}
\end{table}

\begin{figure*}[!ht]
	\subfloat[AHC clustering using different distances and linkage methods]{
        \includegraphics[width=0.5\linewidth]{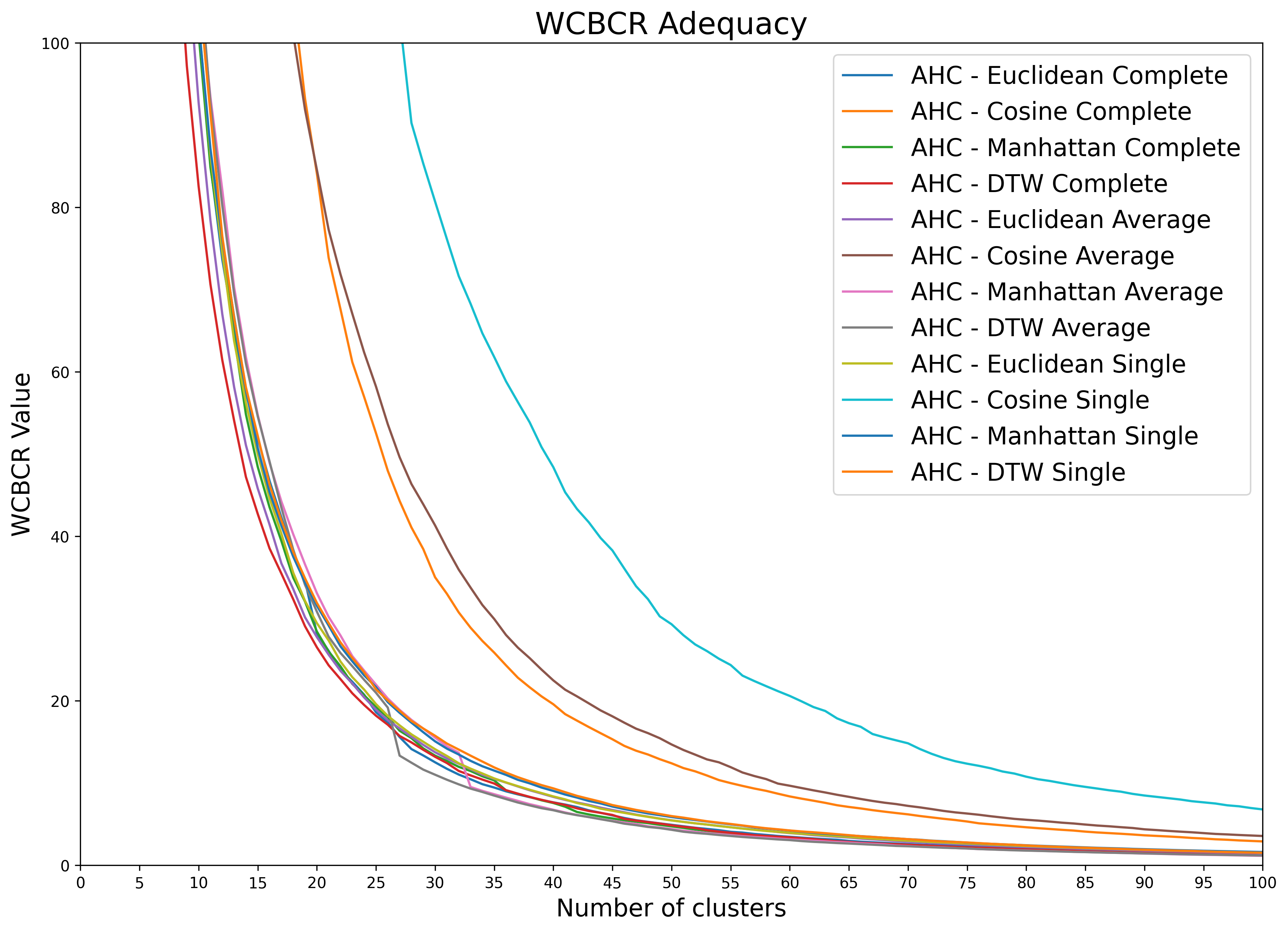}
		\label{fig:ahc-diff-sim-normal}
    }
	\hfill
	\subfloat[AHC clustering with zoom into lower cluster numbers]{
		\includegraphics[width=0.5\linewidth]{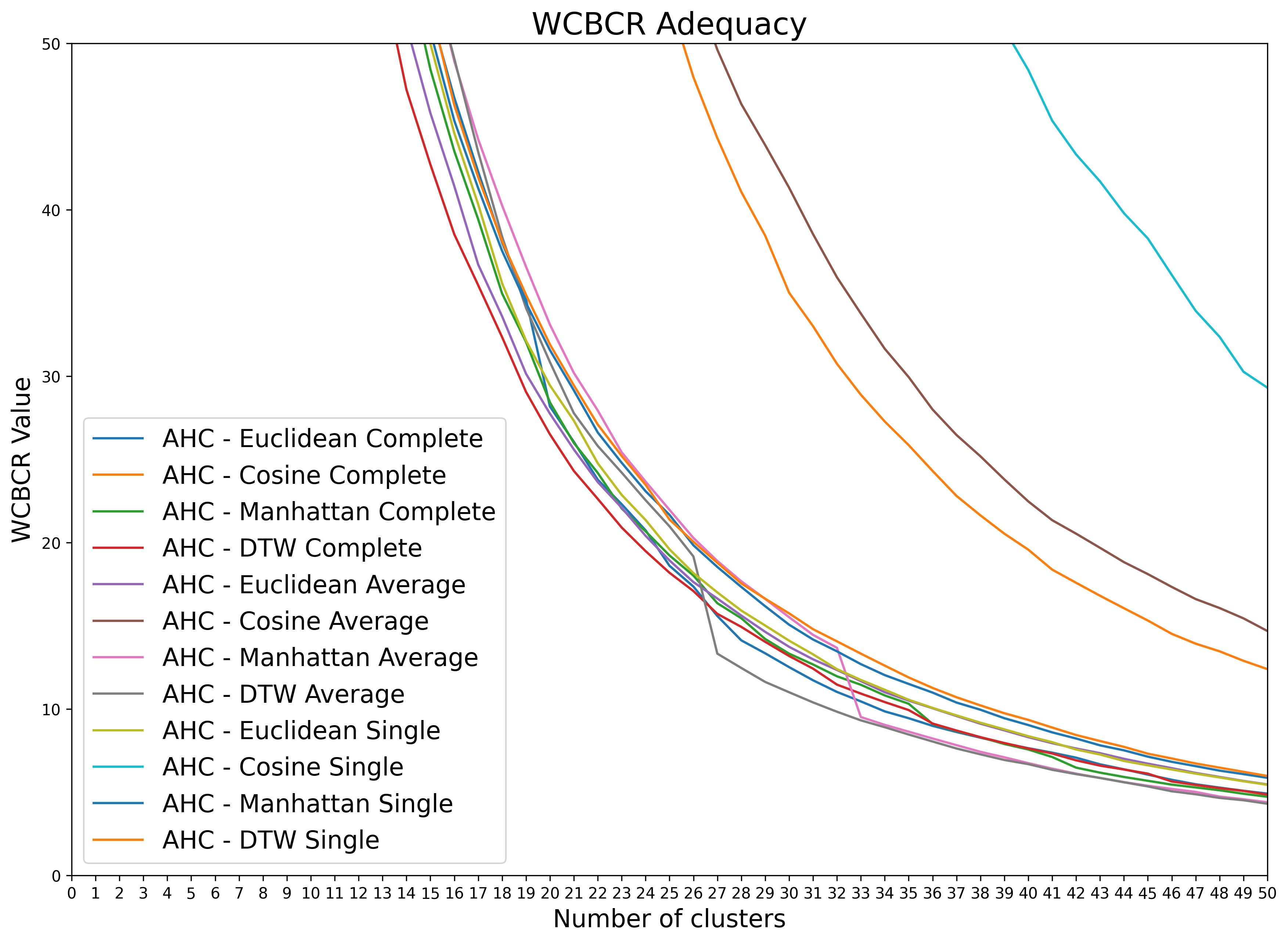}
		\label{fig:ahc-diff-sim-zoom}
	}
	\caption{WCBCR results using AHC with different similarity distances. The figure shows that using AHC with DTW and average linkage criteria outperformed other clustering methods with the optimal number of clusters 27 determined using the elbow method}
	\label{fig:ahc-diff-sim}
\end{figure*}

While Table \ref{tbl:wcbcr-clustering} and Figure \ref{fig:all-clustering-results} compare AHC-DTW with average linkage, Table \ref{tbl:wcbcr-ahc-methods} and Figure \ref{fig:ahc-diff-sim} are strictly using AHC with different similarity distances and linkage criteria. The linkage criteria in Table \ref{tbl:wcbcr-ahc-methods} are labeled as single-linkage \textbf{S}, complete-linkage \textbf{C}, and unweighted-average labeled \textbf{A}. We can conclude that AHC using Complete linkage criteria achieved better WCBCR results when the number of clusters used was less than 23. However, when clusters were more than 26, AHC using the Average Linkage criteria provided better WCBCR values. Furthermore, using the Elbow method, we infer that AHC using Average Linkage requires fewer clusters (27 clusters) with the best WCBCR value of $13.33$.

As seen from Table  \ref{tbl:wcbcr-clustering}, for 27 clusters, AHC with Cosine distance achieved the worst performance irrelevant of which linkage criteria were used. Results obtained with Manhattan and Euclidean distances are close to results of DTW for the same linkage criteria. However, using complete linkage, Euclidean distance achieved a slightly better WCBCR value of 15.59 than DTW's value of 15.72. Nevertheless, the best clustering was achieved with DTW with average linkage.

\begin{table*}[!ht]
    \caption{WCBCR using AHC with different distances and linkage criteria: single-linkage (S), complete-linkage (C), and unweighted-average (A). The green highlighted cell pertains to the best WCBCR $13.33$ from AHC-DWT using average linkage criteria with the optimal number of clusters $27$ determined using the elbow method.
    }
    \centering
    \begin{tabular}{>{\centering}m{0.06\linewidth}
                    >{\centering}m{0.05\linewidth}
                    >{\centering}m{0.05\linewidth}
                    >{\centering}m{0.05\linewidth}
                    >{\centering}m{0.06\linewidth}
                    >{\centering}m{0.06\linewidth}
                    >{\centering}m{0.06\linewidth}
                    >{\centering}m{0.05\linewidth}
                    >{\centering}m{0.05\linewidth}
                    >{\centering}m{0.05\linewidth}
                    >{\centering}m{0.05\linewidth}
                    >{\centering}m{0.05\linewidth}
                    >{\centering}m{0.05\linewidth}}
        \hline\hline
        \rowcolor{lightgray} Clustering \# & DTW S. & DTW C. & DTW A. & Manhattan S. & Manhattan C. & Manhattan A. & Euclidean S. & Euclidean C. & Euclidean A. & Cosine S. & Cosine C. & Cosine A. \tabularnewline
        \hline\hline
        20 & 31.91 & 26.51 & 30.85 & 31.57 & 28.43 & 33.10 & 29.45 & 28.18 & 27.75 & 221.52 & 84.15 & 84.58 \tabularnewline
        21 & 29.45 & 24.31 & 27.78 & 29.15 & 25.99 & 30.20 & 27.33 & 26.06 & 25.60 & 202.14 & 73.86 & 77.27 \tabularnewline
        22 & 27.10 & 22.63 & 25.81 & 26.62 & 24.19 & 27.94 & 24.77 & 23.78 & 23.64 & 165.57 & 67.61 & 71.86 \tabularnewline
        23 & 25.21 & 20.91 & 24.21 & 24.80 & 22.05 & 25.43 & 22.87 & 22.31 & 22.10 & 139.30 & 61.18 & 67.06 \tabularnewline
        24 & 23.48 & 19.49 & 22.55 & 23.09 & 20.68 & 23.67 & 21.36 & 20.74 & 20.41 & 130.07 & 56.95 & 62.36 \tabularnewline
        25 & 21.39 & 18.19 & 20.98 & 21.67 & 19.23 & 21.97 & 19.60 & 18.61 & 18.93 & 121.26 & 52.55 & 58.22 \tabularnewline
        26 & 20.05 & 17.09 & 19.17 & 19.84 & 18.02 & 20.28 & 18.17 & 17.36 & 17.63 & 113.46 & 47.97 & 53.63 \tabularnewline
        \underline{\textbf{27}} & 18.79 & 15.72 & \cellcolor[HTML]{03C03C}\textbf{13.33} & 18.54 & 16.36 & 18.90 & 17.01 & 15.59 & 16.64 & 103.08 & 44.31 & 49.62 \tabularnewline
        28 & 17.55 & 14.94 & 12.47 & 17.34 & 15.46 & 17.68 & 15.92 & 14.13 & 15.62 & 90.27 &  41.08 & 46.36 \tabularnewline
        29 & 16.63 & 14.04 & 11.63 & 16.18 & 14.21 & 16.62 & 15.02 & 13.35 & 14.64 & 85.33 &  38.45 & 43.89 \tabularnewline
        30 & 15.75 & 13.19 & 11.01 & 15.07 & 13.33 & 15.52 & 14.11 & 12.52 & 13.74 & 80.71 &  35.02 & 41.33 \tabularnewline
        \hline
    \end{tabular}
    \label{tbl:wcbcr-ahc-methods}
\end{table*}

\section{Conclusion and Future Work}\label{sec:conclusion}
Energy companies aim to provide an efficient and reliable Demand Response (DR). Therefore, it is imperative to understand and categorize customers' consumption patterns to help energy companies reduce consumption during peak hours by creating incentive programs to target customers based on their consumption patterns. Targeting the right customer group to reduce energy consumption requires placing the customers into categories based on their similarity of consumption patterns. Grouping residential-households load curves are challenging since load curves vary between customers and within the same customer.

This paper proposed AHC with DTW to categorize DLCs based on shape similarity. AHC provides a better starting point for clusters since each cluster prototype is initiated by an actual load curve, compared to K-means and K-medoids, where initial cluster centers are randomly selected. On the other hand, DTW tries to find the optimal alignment between two curves by detecting shifts in the patterns, making DTW a better measurement distance than other distance measures.

All clustering methods quality was evaluated using WCBCR. The optimal number of clusters was determined using the Elbow method at 27, with the best WCBCR value of 13.3 obtained from AHC-DTW using average linkage criteria. Future work will examine the optimal number of clusters using Davies–Bouldin index and evaluate results on a large dataset and perform detailed cluster analysis.

\section*{Acknowledgment}
This research has been supported by NSERC under grant RGPIN-2018-06222.

\bibliographystyle{IEEEtran}
\bibliography{references}

\end{document}